\pdfoutput=1

\documentclass[11pt]{article}

\usepackage{EMNLP2023}

\usepackage{times}
\usepackage{latexsym}
\usepackage{ragged2e}

\usepackage[T1]{fontenc}

\usepackage[utf8]{inputenc}

\usepackage{microtype}

\usepackage{inconsolata}

\usepackage{graphicx}
\usepackage{booktabs} %
\usepackage{tabularx}

\usepackage{amsmath}
\usepackage{amssymb}
\usepackage{mathtools}
\usepackage{amsthm}
\usepackage{amsmath}

\usepackage{lipsum}  
\usepackage{listings}
\usepackage{caption}
\usepackage{subcaption}
\usepackage{CJKutf8}
\usepackage{xspace}
\usepackage{multirow}
\usepackage[linewidth=1pt]{mdframed}
\usepackage{xspace}
\usepackage{enumitem}
\usepackage[subtle]{savetrees}

\definecolor{cb-0}{RGB}{216, 27, 96}
\definecolor{cb-1}{RGB}{30,136,229}
\definecolor{cb-2}{RGB}{255,193,7}
\definecolor{cb-3}{RGB}{0, 77, 64}
\definecolor{cb-4}{RGB}{150,220,174}

\DeclareMathSizes{10}{9}{6}{5}

\NewDocumentCommand{\todo}{ mO{} }{\textcolor{magenta}{\textsuperscript{\textit{TODO}}\textsf{\textbf{\small[#1]}}}}

\NewDocumentCommand{\david}{ mO{} }{\textcolor{red}{\textsuperscript{\textit{David}}\textsf{\textbf{\small[#1]}}}}

\NewDocumentCommand{\suzie}{ mO{} }{\textcolor{blue}{\textsuperscript{\textit{Suzie}}\textsf{\textbf{\small[#1]}}}}
\NewDocumentCommand{\kate}{ mO{} }{\textcolor{green}{\textsuperscript{\textit{Kate}}\textsf{\textbf{\small[#1]}}}}

\newcommand{\Ours}{ALOHa\xspace}
\newcommand{\OursLong}{\underline{A}ssessment with \underline{L}anguage models for \underline{O}bject \underline{Ha}llucination\xspace}
\newcommand{\OurDataset}{HAT\xspace}
\newcommand{\OScore}{ALOHa$_\text{o}$\xspace}

\usepackage{etoolbox}
\makeatletter
\patchcmd{\hyper@makecurrent}{%
    \ifx\Hy@param\Hy@chapterstring
        \let\Hy@param\Hy@chapapp
    \fi
}{%
    \iftoggle{inappendix}{%
        \@checkappendixparam{chapter}%
        \@checkappendixparam{section}%
        \@checkappendixparam{subsection}%
        \@checkappendixparam{subsubsection}%
        \@checkappendixparam{paragraph}%
        \@checkappendixparam{subparagraph}%
    }{}%
}{}{\errmessage{failed to patch}}

\newcommand*{\@checkappendixparam}[1]{%
    \def\@checkappendixparamtmp{#1}%
    \ifx\Hy@param\@checkappendixparamtmp
        \let\Hy@param\Hy@appendixstring
    \fi
}
\makeatletter

\newtoggle{inappendix}
\togglefalse{inappendix}

\apptocmd{\appendix}{\toggletrue{inappendix}}{}{\errmessage{failed to patch}}

\newcolumntype{Y}{>{\centering\arraybackslash}X}

\title{ALOHa: A New Measure for Hallucination in Captioning Models}

\author{Suzanne Petryk\textsuperscript{*}, David M. Chan\thanks{\textsuperscript{*} Indicates equal authorship.}, Anish Kachinthaya, Haodi Zou, \\ \textbf{ John Canny, Joseph E. Gonzalez, Trevor Darrell}\\
        University of California, Berkeley \\ \texttt{\fontsize{11pt}{11pt}\selectfont \{spetryk,davidchan,anishk,haodi.zou,canny,jegonzal,trevordarrell\}@berkeley.edu}\\
        {\fontsize{11pt}{11pt}\selectfont \url{https://davidmchan.github.io/aloha} }}

\begin{document}
\maketitle

\begin{abstract}
    Despite recent advances in multimodal pre-training for visual description, state-of-the-art models still produce captions containing errors, such as hallucinating objects not present in a scene. The existing prominent metric for object hallucination, CHAIR, is limited to a fixed set of MS COCO objects and synonyms. In this work, we propose a modernized open-vocabulary metric, ALOHa, which leverages large language models (LLMs) to measure object hallucinations. Specifically, we use an LLM to extract groundable objects from a candidate caption, measure their semantic similarity to reference objects from captions and object detections, and use Hungarian matching to produce a final hallucination score. We show that ALOHa correctly identifies 13.6\% more hallucinated objects than CHAIR on HAT, a new gold-standard subset of MS COCO Captions annotated for hallucinations, and 30.8\% more on nocaps, where objects extend beyond MS COCO categories.
\end{abstract}

\section{Introduction and Background}
 In recent years, vision-language models have demonstrated remarkable performance. Unfortunately, even state-of-the-art models for visual description still generate captions with object hallucinations -- objects or entities that are present in the caption yet are not explicitly supported by visual evidence in the image \cite{dai2022plausible}. In order to reduce the occurrence of object hallucinations in vision-language models, it is helpful to understand and quantify the problem through \textit{reliable}, \textit{localizable}, and \textit{generalizable} measures of object hallucination. \textit{Reliable} measures are capable of correctly indicating if a given caption contains an object hallucination. \textit{Localizable} measures are capable of indicating which object in a particular caption is hallucinated. \textit{Generalizable} measures are capable of evaluating captions from a wide range of input datasets, across a wide range of object and entity categories.

\begin{figure}
    \centering
    \includegraphics[width=\linewidth]{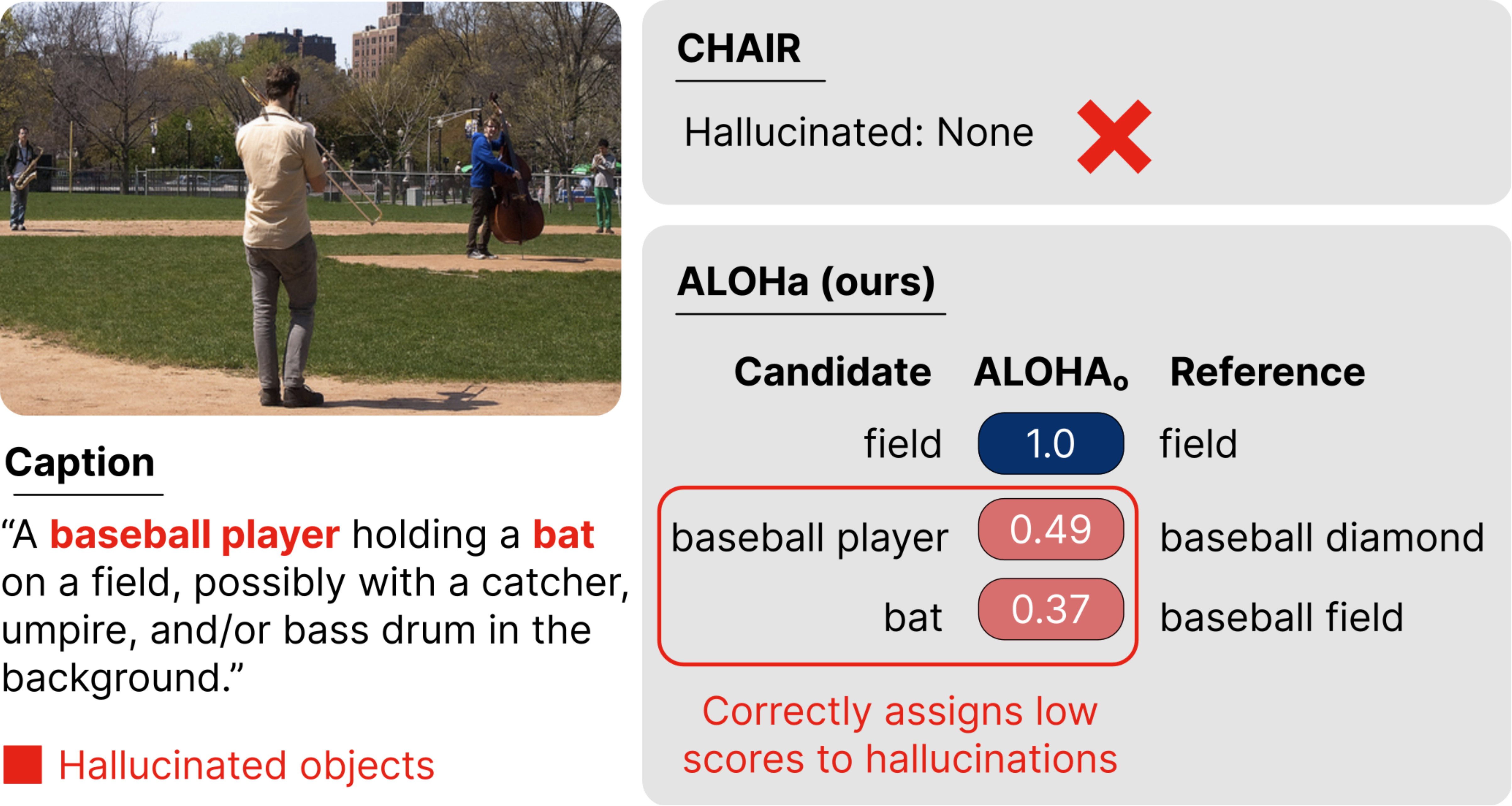}
    \caption{\small (Top) The SOTA prior object hallucination metric, CHAIR, is limited to MS~COCO objects, and fails to detect the hallucinations in this image caption while \Ours (ours, bottom) correctly assigns low similarity scores to the hallucinations ``baseball player'' and ``bat''. \Ours does not penalize the caption for ``catcher'', ``umpire'', and ``bass drum'', as the caption indicates uncertainty of their presence.}
    \label{fig:teaser}
    \vspace{-1em}
\end{figure}

Recent works that measure object hallucinations in generated text generally fall into two categories: measures that find hallucinations by explicitly matching from a set of objects, and measures that compute distances between latent image and/or text embeddings,
indicating a hallucination if the embeddings are too distant. In the first category, CHAIR \cite{rohrbach2018object} is a measure that explicitly extracts objects from candidate sentences using simple string matching against MS COCO classes and a small set of synonyms. It compares these extracted objects against the ground truth detections and objects extracted from the ground truth reference captions. CHAIR is both reliable, as string matching on a fixed set of objects is accurate, consistent, and localizable, as individual non-matching strings are identified. However, as seen in \autoref{fig:teaser}, CHAIR is not generalizable, as it can only handle a fixed set of predetermined objects. Other uni-modal measures in this category include those for abstractive summarization \cite{durmus2020feqa,kryscinski2019evaluating,maynez2020faithfulness,son2022harim+,sridhar2022improved,yuan2021bartscore}, dialogue \cite{huang2022ed,shuster2021retrieval}, and structured knowledge \cite{dhingra2019handling}. These often generalize poorly to vision-language tasks as they require grounding the generated text into inputs of the same modality.

In the second category, CLIPScore \cite{clipscore} employs CLIP \cite{radford2021learning} embeddings to assess image-text matches. While it is generalizable and reliable, it lacks localization as it does not pinpoint incorrect spans of text. CLIPBERTS \cite{wan2022evaluating} and RefCLIPScore (an extension of CLIPScore accounting for reference captions) face similar limitations.

POPE \cite{li2023evaluating} evaluates vision-language models' likelihood to hallucinate objects with machine-generated queries consisting of samples extracted from both reference object detections and nonexistent objects, but addresses a different problem from that which we investigate here -- it measures how often \textit{models} hallucinate rather than localizes and detects issues within \textit{a single caption}.

Inspired by recent successes using LLMs for evaluation in language-only tasks \cite{bert-score, yuan2021bartscore, bubeck2023sparks,chiang2023vicuna,zheng2023judging}, we introduce \OursLong (\Ours), a modernized measure for object hallucination detection that is \textit{reliable}, \textit{localizable}, and \textit{generalizable}. \Ours extends the reliability and localization of CHAIR to new input domains by leveraging in-context learning of LLMs combined with semantically rich text embeddings for object parsing and matching (\autoref{fig:teaser}). 

For a given image caption, we generate two measures: \textbf{\OScore}, a numeric score for each object rating the degree to which that object is a hallucination, and \textbf{\Ours}, an aggregated score rating the degree to which the whole caption contains a hallucination. We demonstrate \Ours on a new gold-standard dataset of image hallucinations, \OurDataset, and show that \Ours improves on CLIPScore while detecting object hallucinations, and CHAIR while correctly localizing those hallucinations. We conclude by demonstrating that \Ours remains reliable and localizable when generalizing to out-of-domain data.

\section{\Ours: Reliable, Localizable, and Generalizable Hallucination Detection}
\label{sec:methods}
\label{sec:method-matching}

\begin{figure*}
    \centering
    \includegraphics[width=\linewidth]{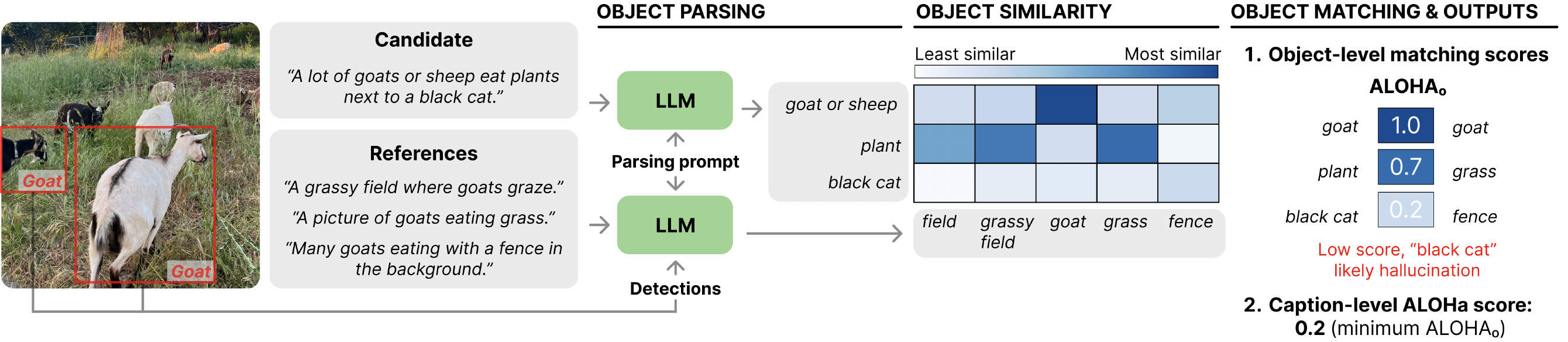}
    \caption{ Overview of \Ours. We prompt an LLM to extract visually grounded nouns from a candidate's machine-generated description and a set of references. We consider uncertain language (e.g., \textit{``goat or sheep''}), add reference objects with and without modifiers (e.g., both \textit{``field''} and \textit{``grassy field''}), and avoid non-visual nouns (e.g., \textit{``picture''} and \textit{``background''}). Then, we compute a maximum-similarity linear assignment between candidate and reference object sets, the weights of which form the \OScore. Matched pairs with low \OScore are likely hallucinations (e.g., \textit{``black cat''}, \OScore$=0.2$). We additionally output the minimum \OScore as a caption-level \Ours score.}
    \label{fig:method}
\end{figure*}

\Ours produces numeric scores rating the degree of hallucination for each object in a candidate caption as well as an overall caption score, given a set of ground-truth reference captions and predicted (or ground truth) image object detections. \Ours consists of three stages (\autoref{fig:method}). (1) Objects are extracted from the image, reference set, and candidate caption using a combination of an object detector and LLM. (2) We filter the object sets and compute semantic representations of each object. (3) We compute a maximum-similarity linear assignment between candidate and reference objects. The scores from each of the pairs in the linear assignment, which we call \OScore, measure the degree of hallucination for each of the candidate objects. The minimum similarity in this linear assignment (the \Ours score) measures the degree of hallucination of the caption.

\noindent\textbf{(1) Extracting objects from candidates, references, and images:} Parsing visually grounded objects in a caption in an open-domain context is a surprisingly difficult task. CHAIR~\citep{rohrbach2018object} relies on a fixed set of MS~COCO objects and synonyms, requiring considerable effort to extend to other datasets, and sometimes failing at ambiguous parses (such as mistaking the adjective ``orange'' for a noun). SPICE~\citep{anderson2016spice} relies on standard grammar-based object parsing, which can have similar issues, as purely text-based methods fall short at identifying which nouns are \textit{visual} -- for instance, avoiding ``picture'' and ``background'' in \autoref{fig:method}. Captions may also indicate uncertainty around object presence, such as ``a bowl or plate'', or ``a dog biting something, possibly a Frisbee.'' We aim to handle such uncertain objects to avoid unfair hallucination penalties.

With the understanding that open-domain parsing is the primary factor in CHAIR's lack of generalization, we leverage the capability of zero-shot in-context learning in large language models. Following \citet{brown2020language}, we use an LLM (ChatGPT,~\citet{chatgpt})  along with the prompt given in \autoref{app:prompt} to turn the parsing task into a language completion task easily solvable by an LLM. We encourage the LLM to extract visual objects in the scene, consisting primarily of noun phrases (including any attributes, such as ``big dog'' and ``purple shirt''), from the candidate and reference captions. We run the LLM against the candidate caption to produce the unfiltered object set $\mathcal{C}$, and again for the corresponding reference captions to produce object set $\mathcal{R}$. To extract objects from the image context, similar to CHAIR, we augment the set of reference objects with objects detected directly from the image using DETR \cite{carion2020end} fine-tuned on MS~COCO.

\noindent\textbf{(2) Object filtering:} We further refine candidate ($\mathcal{C}$) and reference ($\mathcal{R}$) object sets to better reflect specific challenges of object hallucination detection. Ideally, hallucination measures should penalize specificity when candidate attributes are not supported by references (e.g., if ``purple shirt'' $\in \mathcal{C}$, yet ``white shirt'' $\in \mathcal{R}$), but should not penalize generality (e.g., ``shirt'' $\in \mathcal{C}$, yet ``white shirt'' $\in \mathcal{R}$). Thus, we use spaCy~\citep{spacy} to augment $\mathcal{R}$ with the root nouns from each \textit{reference} noun phrase, but leave the candidates unchanged.

Beyond specificity, captions may also express uncertainty about the presence of objects in an image. For conjunctions (e.g., ``fork or knife''), we aim to avoid unfair penalties if at least one of the objects is grounded. \Ours considers all combinations of selecting a single object from each conjunction, denoted as $\mathcal{C}_{\{1\dots M\}}$ and $\mathcal{R}_{\{1\dots N\}}$ (e.g., ``fork'' $\in \mathcal{R}_0$ and ``knife'' $\in \mathcal{R}_1$). Additionally, we prompt the LLM to indicate uncertain grounding by including ``possibly'' after the object (e.g., ``there may be a Frisbee'' becomes ``Frisbee (possibly)'') and we remove uncertain objects from $\mathcal{C}_i$ to avoid penalties while maintaining them in $\mathcal{R}_j$ for maximum coverage of more general objects.

\noindent\textbf{(3) Object Matching:} Once we have extracted and parsed the candidate and reference object sets, we aim to measure the degree of hallucination for each candidate object. While we could match objects based on string alone (resulting in a binary decision), as does CHAIR, often it is useful to understand a continuous scale of hallucination -- e.g., for a reference object ``dog'', hallucinating ``wolf'' should be penalized less than ``potato.'' To capture this scale of semantic similarity, for each object text $o$, we generate $o_{\text{emb}} = \phi(o) \in \mathbb{R}^{K}$, where $\phi$ is a semantic text embedding model. In our work, we use S-BERT \cite{reimers-2019-sentence-bert}. We then compute a similarity score for each pair of objects (usually the cosine similarity, see \autoref{app:similarity}). For each $(\mathcal{C}_i, \mathcal{R}_j)$ pair, we store these scores in a similarity matrix $\mathcal{S}_{i,j} \in [0,1]^{|\mathcal{C}_i| \times |\mathcal{R}_j|}$. We then use the Hungarian method~\citep{kuhn1955hungarian} to find an optimal maximum-similarity assignment $\mathcal{M}_{i,j}$ between candidate and reference sets of objects. 

To determine the \OScore score for each object, we take the maximum score across all possible parsings, giving the candidate caption the benefit of the doubt, for an object $c \in \mathcal{C}_i$
\begin{equation}
    \text{\OScore}(c) = \max_{i, j}\: w_{c_i,j} \in \mathcal{M}_{i,j}
\end{equation}
While $0 \le$ \OScore $\le 1$ indicates the degree of hallucination for each object, we also want to indicate if an entire caption contains a hallucination. We thus define:
\begin{equation}
\label{eq:ours}
\text{\Ours} = \min_{c \in \mathcal{C}} \text{\OScore}(c)
\end{equation}
We choose the minimum as the presence of \textit{any} hallucinated object indicates that the full caption is a hallucination, and even several correct detections should not compensate for a hallucination.

\section{Evaluation \& Discussion}
\label{sec:results}

\begin{table}
\begin{center}
\begin{small}
\begin{tabularx}{\linewidth}{Xcc}
\toprule
Method & LA & AP  \\
\midrule
Baseline (Majority Vote) & - & 33.75 \\
CHAIRs & 6.70  & 36.85 \\
CLIPScore & - & 40.10 \\
RefCLIPScore & - & 48.40 \\ \midrule
\Ours (No Soft Object Matching) & 18.66 & 47.27 \\
\Ours (No Detections) & 19.55 & 48.40 \\
\Ours (Oracle Detections) & 19.55 & 47.86 \\
\Ours (DETR Detections)\textbf{*} & \underline{20.30} & \underline{48.62} \\
\Ours (Oracle$+$DETR Detections) & \textbf{21.05} & \textbf{48.78} \\
\bottomrule
\end{tabularx}
\end{small}
\end{center}
\caption{ Test set performance for binary hallucination detection on \OurDataset. LA: Localization Accuracy. AP: Average Precision. \textbf{*} indicates the version of \Ours used throughout this paper, unless noted otherwise. Oracle detection are human-generated reference detections.}
\label{tab:hat}
\end{table}

\noindent\textbf{\OurDataset:} To promote the development of high-quality methods for hallucination detection, we collect and release \OurDataset (\underline{HA}llucination \underline{T}est), a dataset of labeled hallucinations in captions. \OurDataset consists of 490 samples (90 validation and 400 test) labeled by in-domain experts for hallucination on both a word level and caption level (See \autoref{app:dataset}). Measures are evaluated on two metrics: Average Precision (AP) and Localization Accuracy (LA). The AP of the method measures reliability and is defined as how well the measure identifies captions with hallucinations. For CHAIR, decisions are binary, so $\text{AP} = \text{accuracy}$. For \Ours, AP is the weighted mean of precisions across all thresholds. The LA, measured on samples containing hallucinations in HAT, measures localization and is defined as the accuracy of correctly indicating \textit{which} of the specific objects were hallucinated. For CHAIR, a hallucination is correctly localized when at least one detected string mismatch is a hallucination, and for \Ours when the minimum \OScore score corresponds to a hallucinated object.

\begin{figure}
    \centering
    \includegraphics[width=\linewidth]{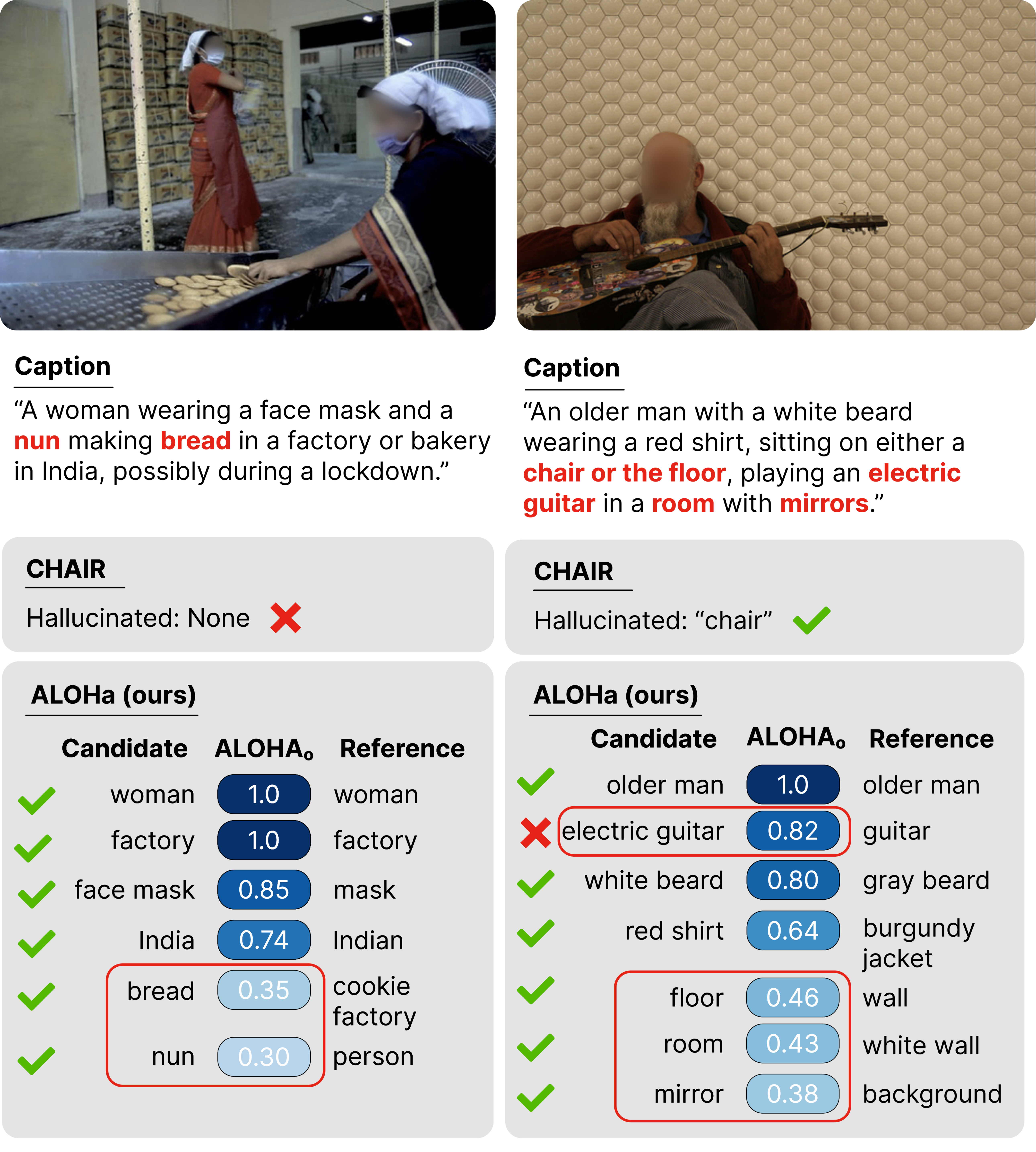}
    \caption{
    Qualitative Flickr30k examples. (Left) \Ours correctly assigns low scores to the hallucinated ``nun'' and ``bread'', whereas CHAIR does not detect any hallucinations. (Right) Although \Ours assigns high similarity between the hallucinated ``electric guitar'' and reference ``(acoustic) guitar'', it assigns low scores to the other 3 hallucinations. CHAIR detects only the hallucination ``chair'', missing the others.} 
    \label{fig:qual-flickr}
\end{figure}

\Ours's performance on \OurDataset is shown in \autoref{tab:hat}. On AP, \Ours with DETR detections outperforms both CHAIR and CLIPScore by 11.8\% and 8.5\% respectively. RefCLIPScore attains a similar AP; however, is not localizable. \Ours achieves more than twice the LA on \OurDataset CHAIR, a particularly challenging task as \OurDataset includes non-object hallucinations, such as incorrect verbs or relations (see \autoref{fig:app-hat-random-hc}). \autoref{tab:hat} further ablates the choice of image detections and indicates that \Ours is robust to missing detections.

\begin{table*}
\begin{center}
\begin{small}
\begin{tabularx}{\linewidth}{Xcccccccccc}
\toprule
& \multicolumn{2}{c}{FOIL} & \multicolumn{8}{c}{nocaps-FOIL} \\
\cmidrule(l{8pt}r{8pt}){2-3} \cmidrule(l{8pt}r{8pt}){4-11}
& \multicolumn{2}{c}{Overall} & \multicolumn{2}{c}{In-Domain} & \multicolumn{2}{c}{Near-Domain} & \multicolumn{2}{c}{Out-Domain} & \multicolumn{2}{c}{Overall} \\
\cmidrule(l{8pt}r{8pt}){2-3} \cmidrule(l{8pt}r{8pt}){4-5} \cmidrule(l{8pt}r{8pt}){6-7} \cmidrule(l{8pt}r{8pt}){8-9} \cmidrule(l{8pt}r{8pt}){10-11}
Method & LA & AP & LA & AP & LA & AP & LA & AP & LA & AP \\
\midrule

Baseline (Majority Vote) & - & 50.00 & - & 50.00 & - & 50.00 & - & 50.00 & - & 50.00 \\
CHAIRs & \textbf{79.00} & \textbf{92.50} & 13.47 & 57.82 & 17.55 & 59.14 & 12.24 & 58.06 & 14.42 & 58.33 \\
CLIPScore & - & 76.44 & - & \underline{71.81} &  - & \underline{70.17} & - &  \underline{78.73} & - & \underline{73.48} \\
RefCLIPScore & - & \underline{80.64} & - & \textbf{79.63} & - & \textbf{78.70} & - & \textbf{85.89} & - & \textbf{81.31} \\
\midrule
\Ours & 40.00 & 61.35 & \textbf{47.35} & 71.80 & \textbf{47.30} & 66.67 & \textbf{48.84} & 70.91 & \textbf{45.17} & 69.52  \\
\bottomrule
\end{tabularx}
\end{small}
\end{center}
\caption{Breakdown of results by domain on FOIL and nocaps FOIL. AP: Average Precision. LA: Localization Accuracy. Bold and underlined values represent the best and second-best methods respectively.}
\label{tab:foil-supp}
\end{table*}

\noindent \textbf{FOIL object hallucinations:} To indicate generalizability we evaluate our method on two machine-generated object hallucination datasets. FOIL \cite{shekhar2017foil} contains MS~COCO images, where objects are randomly replaced with similar ones (e.g., ``bus`` and ``car''), and nocaps-FOIL, a similar dataset that we construct on the nocaps dataset~\citep{agrawal2019nocaps} for novel object captioning beyond MS~COCO (see~\autoref{app:nocaps-foil}). %
\autoref{tab:foil-supp} breaks down the results of \Ours on the FOIL and nocaps-FOIL dataset. The results illustrate a subtle result. While \Ours under-performs CHAIRs in both AP and LA on the original FOIL dataset, this is because FOIL constructs new samples by replacing string-matched COCO objects with a set of hand-selected ``foil'' objects (near semantic neighbors). This is a best-case scenario for CHAIR, as CHAIR relies on fixed object-set string matching alone, and thus, is easily able to both detect and localize the replaced samples. When we move to nocaps-FOIL with non-MS~COCO data, however, \Ours significantly outperforms CHAIR, as now the object set that was a strength for in-domain FOIL becomes a liability, and CHAIR is unable to detect any hallucinations at all, due to the restricted string matching. RefCLIPScore, while competitive in the hallucination detection task, cannot perform localization.

\noindent\textbf{Qualitative Examples - Flickr30k:} In \autoref{fig:qual-flickr} and \autoref{fig:app-qual-flickr}, we visualize the behavior of CHAIR and \Ours on several Flickr30k samples~\cite{young2014image}, using captions generated by a recent captioning model~\citep{chan2023ic} that often produces complex captions with phrases expressing uncertainty.

\noindent \textbf{Ablation - Choice of LLM:} The language model is critical to the overall performance of \Ours - language models with insufficient zero-shot parsing capability will suffer reduced downstream performance. We investigate the performance of the language model in \autoref{tab:llm} on \OurDataset. In addition to LA and AP, we also measure ``Parsing error rate" (PER), which is the rate of errors made when parsing objects from reference captions on \OurDataset, and ``Parsing recall rate (PRR), which is the recall rate of objects in the captions (See \autoref{app:metrics}). 

\begin{figure}[t]
    \centering
    \includegraphics[width=\linewidth]{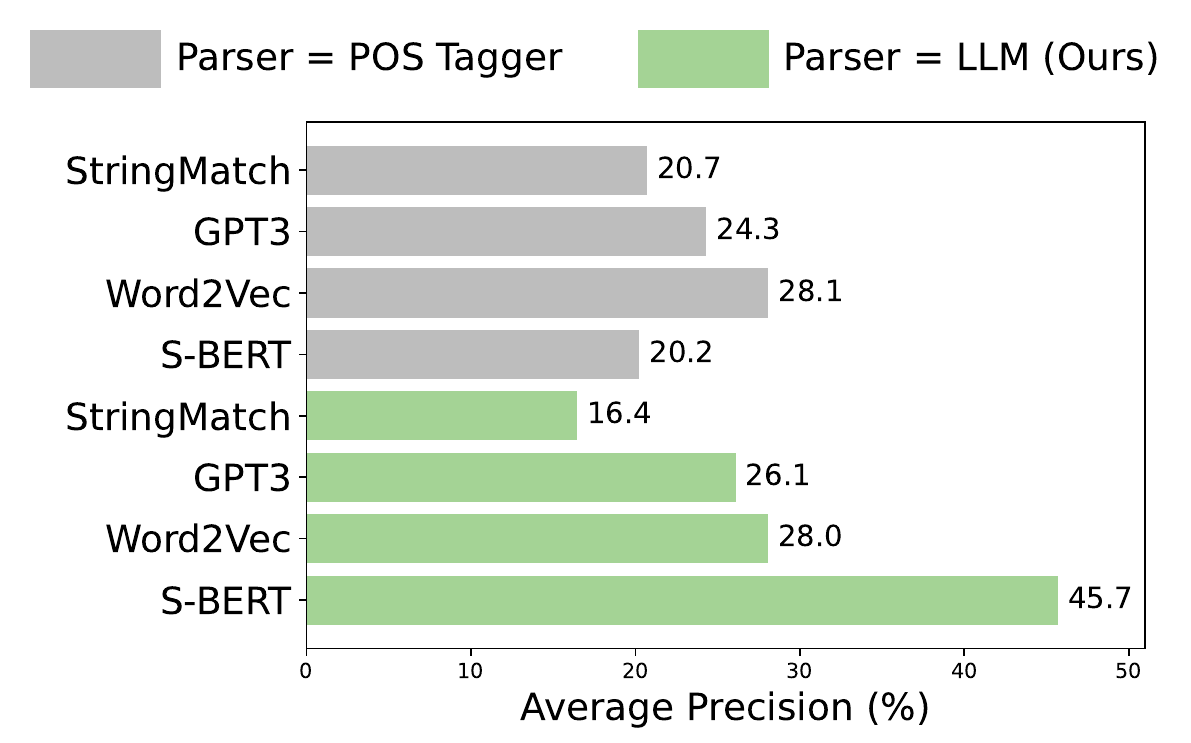}
    \caption{ Performance on \OurDataset validation set filtered for hallucinated objects, when comparing embedding methods and object extraction approaches.}
    \label{fig:results-similarity}
\end{figure}

\noindent \textbf{Ablation - Object Extraction and Semantic Embedding Methods:} In the this work, we leverage LLMs \cite{openai2023gpt4} for object extraction, and a BERT-based model \cite{reimers-2019-sentence-bert} for semantic word embedding. In \autoref{fig:results-similarity}, we explore the difference in overall performance on \OurDataset's validation set when using different combinations of object extraction and semantic embedding. Namely, we compare LLM-based extraction to the parse-tree-based noun extraction in SpaCy \cite{honnibal2020spacy}, and compare SentenceTransformer (BERT-Based model, \cite{reimers-2019-sentence-bert}) to Word2Vec \cite{mikolov2017advances}, GPT-3 (Ada) embedding, and CHAIR-style string matching (following CHAIR, strings are case-normalized and lemmatized). Combining LLMs with the SentenceTransformer (BERT-Based) model outperformed other methods, and fuzzy embedding methods outperformed exact string matching. This is generally expected: humans have a wide vocabulary that is poorly captured by exact string matching. Word2Vec outperforms GPT-3 embeddings. We believe that this is because the GPT-3 embeddings are optimized for sentence-level structures, and may fail to semantically embed single words in a meaningful way.  Interestingly, S-BERT is not a word similarity measure and was instead designed to measure distances between sentences (and could lead to inaccurate single-word judgments) -- While we did find S-BERT most effective among our approaches, we believe that leveraging a large-scale model trained specifically for semantic similarity between words would be an exciting and powerful extension to the \Ours framework.

\begin{table}

\begin{center}
\begin{small}
\begin{tabularx}{\linewidth}{Xcccc}
\toprule
Lanugage Model & LA $\uparrow$ & AP $\uparrow$ & PER $\downarrow$ & PRR $\uparrow$ \\
\midrule
GPT-3.5 & 20.30 & \textbf{48.62} & \textbf{2.97} & \textbf{98.63} \\
Claude (Instant) & \underline{20.74} & \underline{41.48} & \underline{3.31} & - \\
Koala & \textbf{22.22} & 38.70 & 5.07 & - \\
\bottomrule
\end{tabularx}
\end{small}
\end{center}
\caption{Exploration of LLM choice for parsing within \Ours, on \OurDataset. AP: Average Precision, LA: Localization Accuracy, PER: Parsing Error Rate (\%), PRR: Parsing Recall Rate.}
\label{tab:llm}
\end{table}

\section{Conclusion}
This paper introduces \Ours, a scalable LLM-augmented metric for open-vocabulary object hallucination. \Ours correctly identifies 13.6\% more hallucinated objects on \OurDataset and 31\% on nocaps-FOIL than CHAIR. \Ours represents an important modernization of caption hallucination metrics, and detecting complex hallucinations in actions, quantities, and abstract concepts remains an exciting and challenging task for future exploration.

\section*{Limitations / Ethical Considerations}
\label{sec:limits}

While \Ours represents a strong step towards open-domain localized hallucination detection, it comes with several limitations which we discuss in this section.

\paragraph{Non-determinism}
A primary concern with using large language models for an evaluation measure is the natural nondeterminism that comes with them. While in theory language models sampled at a temperature of zero (as we do in this work) are deterministic, it is well documented that small random fluctuations can still occur \cite{openai2023gpt4}. Beyond random fluctuations, the availability of language models long-term can impact the reproducibility of the measure. In this work, we primarily rely on closed-source language models, which can change or become unavailable without notice. In \autoref{tab:llm}, we demonstrate that \Ours still functions with open source models such as Koala \cite{koala_blogpost_2023}, however, the performance is significantly degraded due to the parsing capabilities of the model. With time, and more powerful open-source LLMs, this will become less of an issue, however relying on a nondeterministic metric for comparative evaluation can easily become a liability.

\paragraph{Availability of Reference Captions (Reference-Free vs. Reference-Based Measures)}
One of the primary limitations of the \Ours evaluation method is the requirement that reference captions are available for the evaluation dataset (an issue shared by CHAIR). Not only must reference captions be available, but they also must sufficiently cover the salient details in the reference image. When the references are impoverished (as can easily happen with a single reference sentence \cite{chan2023ic}) or when there are no references, and ALOHa must rely entirely on detections, the method under-performs more general methods such as CLIPScore which are reference-free, and rely on a large pre-training dataset to encode vision and language correspondences. We strongly believe that the area of reference-free localized hallucination detection is an important area of future research; how can we leverage the tools from large vision and language pre-training in a localized way to understand and interpret where hallucinations lie in the hallucinated text? That being said, there is also a place for reference-based measures, as reference-based measures focus on what \textit{humans} believe to be salient details in the image, whereas reference-free measures always rely on downstream models which \textit{approximate} what humans believe to be important. This means that reference-based measures can often transfer better to new domains than reference-free measures, which often must be trained/fine-tuned in-domain with human-labeled data to achieve strong performance.

\paragraph{General costs associated with LLMs}
The use of large language models for any task incurs significant compute, monetary, environmental, and human costs. \Ours is a significantly slower evaluation measure than methods like CHAIR (however not that much less efficient than CLIPScore), leading to increased power consumption, and cost during evaluation. In addition, the models that we rely on are generally closed source and represent a non-trivial monetary expenditure (Experiments in this paper, including ablations, testing, and prototyping required approximately USD \$120 in API fees). Such factors can be limiting to researchers who wish to evaluate large datasets, however we hope that with the advent of larger open-source models, and continued investment in hardware and systems research, the cost will decrease significantly. Beyond compute and financial costs, there are environmental and human costs associated with using large language models for evaluation, see \citet{bender2021dangers} for a detailed discussion of these factors.

\paragraph{Limited Control of Bias}
In this work, we do not evaluate the performance of \Ours on Non-English data, nor do we explicitly control for or measure bias in the creation of \OurDataset (Which is a labeled subset, randomly selected of the MS COCO dataset), or the Nocaps-FOIL dataset (which operates on the same samples as the Nocaps validation dataset). While \OurDataset is a subset of the common MS COCO dataset, we recognize that the creation of such potentially biased datasets has the potential to lead researchers to engineer features and methods which are unintentionally biased against underrepresented groups. We aim to address these shortcomings in the next iteration of \OurDataset, which will not only contain out-of-domain data for MS COCO-trained models but also aims to better control for bias in the underlying image and caption data. Note that our work, including \OurDataset, is intended for research purposes.

\section*{Acknowledgements}

We thank Dr. Kate Saenko for their helpful comments on the work. Authors, as part of their affiliation with UC Berkeley, were supported in part by the NSF, DoD, and/or the Berkeley Artificial Intelligence Research (BAIR) industrial alliance program, as well as gifts from Anyscale, Astronomer, Google, IBM, Intel, Lacework, Microsoft, Mohamed Bin Zayed University of Artificial Intelligence, Samsung SDS, Uber, and VMware.
\bibliography{egbib}
\bibliographystyle{acl_natbib}

\appendix

\appendix

\setcounter{table}{0}
\renewcommand{\thetable}{A\arabic{table}}

\setcounter{figure}{0}
\renewcommand{\thefigure}{A\arabic{figure}}

\section*{Appendix}

\begin{description}
    \item[\autoref{app:prompt}] describes the prompt of the language model, including the exact language used, the design choices, and the in-context examples.
    \item[\autoref{app:experiments}] contains additional experimental details for experiments in the paper.
    \item[\autoref{app:dataset}] describes the datasets that we collected and constructed, including \OurDataset and nocaps-FOIL.
\end{description}

\section{Prompt}
\label{app:prompt}

The choice of prompt for a large language model using in-context learning is critical to the performance of the model. Each component of the prompt has some ability to shape the downstream language distribution. In this work, we use the prompt shown in \autoref{fig:prompt}. This prompt has several rules, which we discuss here.

\begin{figure}[t]
\noindent\begin{minipage}{\linewidth}

\begin{mdframed}
{\RaggedRight%
\texttt{You are an assistant that parses visually present objects from an image caption. Given an image caption, you list ALL the objects visually present in the image or photo described by the captions. Strictly abide by the following rules:} \\

\texttt{- Include all attributes and adjectives that describe the object, if present} \\
\texttt{- Do not repeat objects} \\
\texttt{- Do not include objects that are mentioned but have no visual presence in the image, such as light, sound, or emotions} \\
\texttt{- If the caption is uncertain about an object, YOU MUST include '(possibly)' after the object} \\
\texttt{- If the caption thinks an object can be one of several things, include 'or' and all the possible objects} \\
\texttt{- Always give the singular form of the object, even if the caption uses the plural form}
}
\end{mdframed}
\end{minipage}
\caption{The prompt that we use for parsing objects from both captions and sets of reference captions.}
\label{fig:prompt}
\end{figure}

\paragraph{Attributes:} We ask that the language model include all attributes attached to the object if they are present. By doing so, we can catch hallucinations such as those shown in \autoref{fig:qual-flickr}, where ``electric guitar" appears in the candidate, but an acoustic guitar is shown in the image. Attributes are handled differently between reference captions and candidate captions. For reference captions, we add both the object with attributes, and the object without attributes to the set, so the candidate is not penalized for being more general. For the candidate, however, we add only the object with attributes, so if the candidate produces attributes, they must match with something in the reference set.

\paragraph{Repeated Objects:} In this work, our primary goal is to determine if a particular object is hallucinated, and not focus on the quantity of hallucinations. Thus, we de-duplicate the object set in both the candidate and reference captions, as well as detections coming from the image. By doing this, we focus on whether the objects can exist in the image, rather than focus on getting the exact count, which may be incorrect if a candidate caption mentions the same object more than once (and that object is parsed twice).

\paragraph{Intangible Object:} In many cases, objects mentioned in the candidate or reference set may be intangible, such as color, light, sound, or emotion. To improve the accuracy of the model, we explicitly suggest that such objects should not be included. 

\paragraph{Or/Possibly:} Modern captioning methods such as Chat-Captioner \cite{zhu2023chatgpt} and IC3 \cite{chan2023ic} are capable of encoding uncertainty into their approach through the use of words like ``possibly" or ``maybe". Additionally, they may make judgments that are uncertain such as ``an apple or an orange." Existing captioning and hallucination detection measures fail to account for this uncertainty, and match both objects, even though the semantics of the caption suggests that the object is uncertain, or may be one of many objects. To account for this, we encourage the LLM to indicate uncertainty in a fixed way, as well as list multiple alternatives on a single line. We then account for this in our matching method, by giving the candidate the benefit of the doubt, scoring only the best match from an alternative set, and ignoring any uncertainty. 

\paragraph{Singularization:} While it is possible to singularize objects using rule-based methods, rule-based methods struggle with challenging nouns, and we found that in general, the LLM was better at performing the singularization set of the post-processing before object matching.

\subsection{In-Context Examples}

In addition to the core prompt text, we provide several contextual samples, which help with in-context learning \cite{brown2020language}. The contextual samples help to align the label space of the model correctly with the target output distribution \cite{min2021metaicl}. An example of such contexts is given in \autoref{fig:context1} and \autoref{fig:context2}. 

\begin{figure}[t]
\noindent\begin{minipage}{\linewidth}
\begin{mdframed}
{\RaggedRight%
\texttt{Caption: This image shows two pink roses in a tulip-shaped vase on a wooden kitchen counter, next to a microwave and a toaster oven. } \\

\texttt{Objects: } \\
\texttt{- pink rose } \\
\texttt{- tulip-shaped vase } \\
\texttt{- wooden kitchen counter } \\
\texttt{- microwave } \\
\texttt{- toaster oven }
}
\end{mdframed}
\end{minipage}
\caption{An example of a single-caption parsing result.}
\label{fig:context1}
\end{figure}

\begin{figure}[t]
\noindent\begin{minipage}{\linewidth}
\centering
\begin{mdframed}
{\RaggedRight%
\texttt{Captions:} \\
\texttt{- Several people riding on a motorcycle with an umbrella open.} \\
\texttt{- Couples riding motorcycles carrying umbrellas and people sitting at tables.} \\
\texttt{- A group of people riding scooters while holding umbrellas.} \\
\texttt{- Some tables and umbrellas sitting next to a building.} \\
\texttt{- Pedestrians and motorcyclists near an open outdoor market.} \\

\texttt{Objects:} \\
\texttt{- person} \\
\texttt{- couple} \\
\texttt{- motorcycle} \\
\texttt{- umbrella} \\
\texttt{- table} \\
\texttt{- scooter} \\
\texttt{- building} \\
\texttt{- pedestrian} \\
\texttt{- motorcyclist} \\
\texttt{- open outdoor market} \\
}
\end{mdframed}
\end{minipage}
\caption{An example of a multi-caption parsing result.}
\label{fig:context2}
\end{figure}

\section{Experimental Details \& Additional Experimentation}
\label{app:experiments}

\subsection{Metrics}
\label{app:metrics}

We employ several measures in the paper, which we describe in detail here.

\paragraph{Average Precision} We measure the \textbf{Average Precision (AP)} of each hallucination metric to detect sentence-level hallucinations. Specifically, we label each sample with \textbf{1} if it contains a hallucination and \textbf{0} otherwise. We then measure AP between those labels and per-sample hallucination measures. For \Ours, this is:
\begin{equation}
    \text{AP} = \frac{1}{N}\sum_{i=1}^N \mathbb{I}[\text{label}] \cdot (1 - \text{\Ours})(i)
\end{equation}
For CHAIR, this is:
\begin{equation}
    \text{AP} = \frac{1}{N}\sum_{i=1}^N \mathbb{I}[\text{label}] \cdot \mathbb{I}[\text{CHAIR Prediction}]
\end{equation}

It is worth noting that when computing average precision, we define the positive label (1) to be “hallucination” to measure the ability of ALOHa or CHAIR to correctly identify hallucinations. Indeed, a lower ALOHa indicates that a caption is more likely to have a hallucination – therefore, we negate the ALOHa score when computing AP. We follow the standard method of computing AP with binary labels and continuous confidence values, where precision and recall are iteratively computed with each confidence value (-ALOHa) as the threshold. The AP is an average of those precisions, each weighted by the increase in recall from the previous threshold.

\paragraph{Localization Accuracy} Localization accuracy (LA)  measures the fraction of samples where a metric can correctly identify a hallucinated object, among samples that are known to contain hallucinated objects. 

\begin{equation}
    \text{LA} = \frac{|\{\text{ $\geq 1$ correctly identified halluc.} \}| }{|\{ \text{$\geq 1$ halluc.}\}|}
\end{equation}

\noindent A sample receives LA of 1 if at least one of the predicted hallucinated objects was correct (for CHAIR), or if the object with the minimum matching score was a true hallucination (for \Ours). We do not measure LA for CLIPScores, as they cannot provide hallucination scores per object.

\subsection{Semantic Similarity Measure}
\label{app:similarity}

In \Ours, we compute the similarity between objects using the cosine distance between embedding vectors generated using the \texttt{all-MiniLM-L6-v2}  S-BERT  implementation in the Sentence-Transformers\footnote{\url{https://www.sbert.net/}} library \cite{reimers-2019-sentence-bert}. While in theory cosine distances should lie in the interval $[-1, 1]$, in this library, for optimization stability, models are trained with positive samples having similarity $1$, and negative samples having similarity $0$. This (unintentionally) induces a model which (by optimization) only produces positive cosine similarity scores. \Ours can still be adapted to negative similarity: our algorithms for maximal assignment and equations 1 and 2 both support negative values (even though they don't appear in this instantiation of the algorithm).

\paragraph{Parsing Error Rate (PER) and Parsing Recall Rate (PRR)} We calculate PER (Parsing Error Rate) with manual annotation by taking the fraction of objects output by the LLM that did not exist in the caption (in other words, measuring 1-precision of parsed objects). We additionally annotate and compute the Parsing Recall Rate (PRR) - the fraction of objects in the caption that are included in the objects parsed by the LLM. This gives a recall for GPT-3.5 of 98.63\%. In these experiments, we find that while Koala \cite{koala_blogpost_2023} has strong LA performance on \OurDataset, however ChatGPT (GPT-3.5) \cite{openai2023gpt4} has both the best average precision, and makes the fewest errors, thus we leverage GPT-3.5 for our primary experiments in the main paper. 

\section{Datasets}
\label{app:dataset}

In this section, we discuss further the data that we use and go into detail on the dataset collection process for \OurDataset (\autoref{app:hat}) and the nocaps-FOIL dataset (\autoref{app:nocaps-foil})

\subsection{nocaps-FOIL}
\label{app:nocaps-foil}

The FOIL dataset \cite{shekhar2017foil} is a synthetic hallucination dataset based on samples from the MS-COCO \cite{xu2016msr} dataset. In this dataset, for each candidate-image pair, a ``foil" caption is created which swaps one of the objects (in the MS-COCO detection set) in the caption with a different, and closely related neighbor (chosen by hand to closely match, but be visually distinct). While the FOIL dataset provides a useful benchmark for many hallucination detection methods, it is overly biased towards methods optimized for the MS-COCO dataset. To help evaluate more general methods, we introduce a new dataset ``nocaps-FOIL" based on the nocaps \cite{agrawal2019nocaps} dataset. The nocaps dataset consists of images from the OpenImages \cite{kuznetsova2020open} dataset annotated with image captions in a similar style to MS-COCO. nocaps is split into three sets: an in-domain set, where objects in the images are in the MS-COCO object set, near-domain, where the objects in the image are related to those of MS-COCO, and out-of-domain, where objects in the image are not contained in MS-COCO.

To build the nocaps-FOIL dataset, for each image, we generate the baseline caption by removing a single caption from the reference set. We then generate the foil caption as follows. First, we find any words in the baseline caption that are contained in either the openimages class list (there are 600) or a near neighbor in Wordnet. We then randomly select one of these classes to replace. Because there are 600 classes, we do not hand-pick the foil classes, and rather, select a near neighbor class based on sentence embeddings from \cite{reimers-2019-sentence-bert}. We find that in practice, the nearest neighbor is often a synonym, thus, to avoid selecting synonyms, we take the 10th furthest sample, which is often a near neighbor, but is visually distinct. We replace this word in the caption, matching case, and then perform a filter for grammatical correctness using the Ginger\footnote{\url{https://www.gingersoftware.com/}} API. Any captions which are not grammatically correct are filtered. This leaves us with 2500 image/caption/foil pairs, which we use for evaluation in \autoref{tab:foil-supp}. 

The OpenImages dataset annotations are under a CC BY 4.0 license, and the images are under a CC BY 2.0 license.

\subsection{\OurDataset}
\label{app:hat}

\begin{figure*}
    \centering
    \includegraphics[height=0.8\paperheight,keepaspectratio]{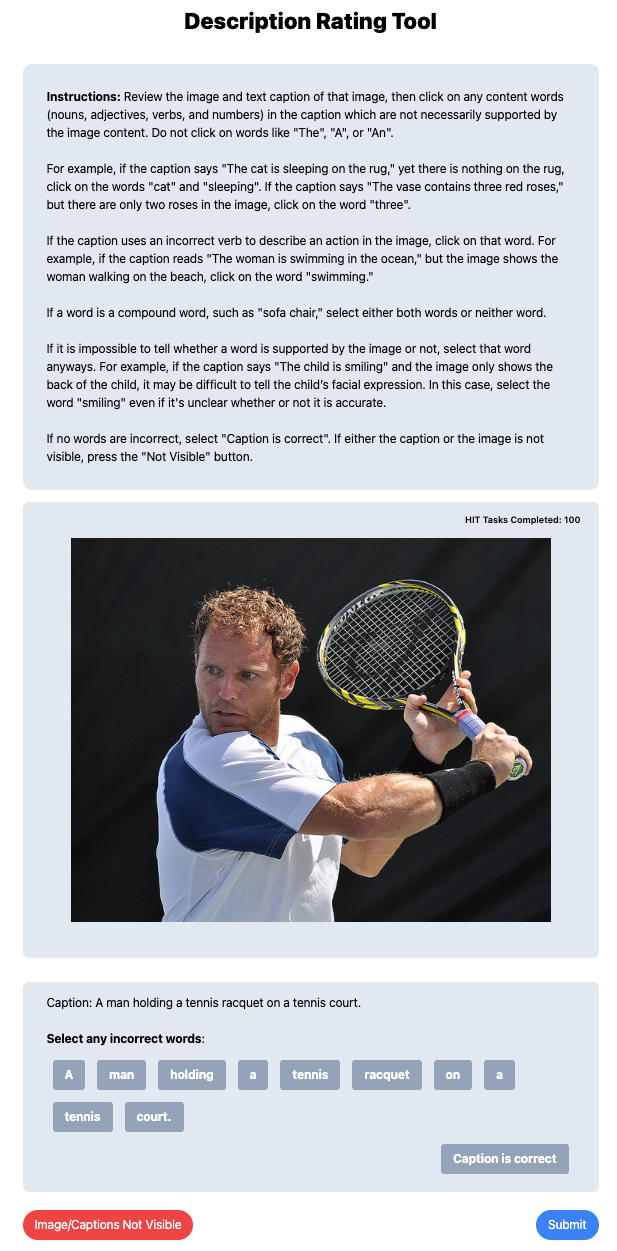}
    \caption{The hallucination dataset collection interface.}
    \label{fig:interface}
\end{figure*}

\OurDataset is based on MS-COCO and aims to be a gold-standard benchmark for the evaluation of hallucination in image captioning methods. While it is relatively small, it is densely annotated by in-domain experts for several types of hallucination including object hallucination, action hallucination, and numeric hallucination among others. \OurDataset consists of 90 validation samples, and 400 test samples, each containing a machine candidate caption generated by one of BLIP \cite{li2022blip}, OFA \cite{wang2022ofa}, IC3 \cite{chan2023ic} or Chat-Captioner \cite{zhu2023chatgpt}, and annotations which mark which word in the captions are hallucinated (See \autoref{fig:interface} for exact instructions given to annotators). An image/caption pair is considered a hallucination if at least one of the words in the caption is hallucinated. 

Screenshots of the interface for data collection are given in \autoref{fig:interface}. While initial versions of the dataset were collected using AMT workers, we found that the quality of annotations was not sufficiently high, and thus, trained experts explicitly in hallucination detection, and leveraged expert ratings for the samples in the test dataset. 

MS-COCO is under a Creative Commons Attribution 4.0 License.

\section{Qualitative Examples}
\label{app:qual}

We provide additional qualitative examples from the following scenarios:

\subsection{Flickr30k Examples}
\label{app:qual-flickr}

\begin{figure*}
    \centering
    \includegraphics[width=\linewidth]{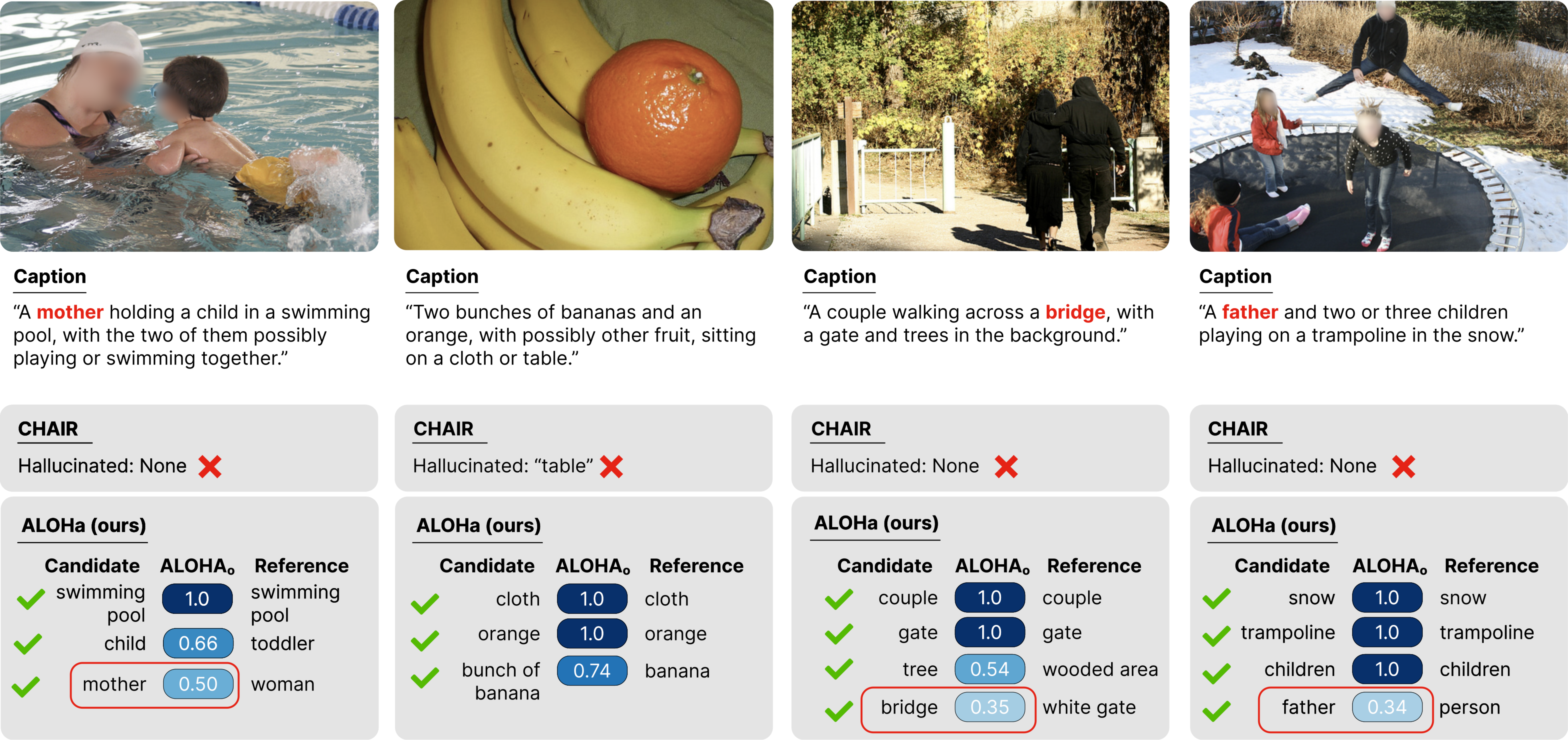}
    \caption{Qualitative samples of \Ours evaluated on the Flickr-30k dataset, with candidate captions generated by IC3 \cite{chan2023ic}. Hallucinated objects in the caption text are red and bolded. See \autoref{app:qual-flickr} for discussion.
    }
    \label{fig:app-qual-flickr}
\end{figure*}

\autoref{fig:app-qual-flickr} shows several examples on the Flickr-30k dataset~\citet{young2014image} with captions generated by IC3 \cite{chan2023ic}, a modern image captioning model that often generates longer, more complex captions including uncertain language such as ``possibly.'' We highlight objects with \OScore $\le 0.5$ as likely hallucinations. For samples going from left to right:

{\RaggedRight
\begin{enumerate}
    \item The caption hallucinates the word ``mother'', as there is no visual evidence that the woman is specifically a mother. CHAIR does not capture this, as ``mother'' is mapped to a synonym for ``person'', which it counts as a grounded (non-hallucinated) object. \Ours matches ``mother'' to the reference ``person'', assigning a borderline \OScore of 0.5. \\
    \item The image does not contain a hallucination. CHAIR flags ``table'' as hallucinated, yet the caption expressed uncertainty with a conjunction: ``chair or table.'' \Ours successfully parses this conjuction and selects ``cloth'' with \OScore $=1.0$ to the exact reference match.\\
    \item CHAIR does not detect the hallucinated ``bridge'', which is successfully assigned a low \OScore$=0.35$.\\
    \item The caption hallucinates the word ``father''. In most cases, the specific relationship of ``father'' is unlikely to be grounded (similar to ``mother'' in sample 1); yet, in this image, it is even more clear as there are only children present. CHAIR maps ``father'' as another synonym for ``person'' and does not consider it a hallucination, whereas ``father'' has a low \OScore$=0.34$.
\end{enumerate}
}

\subsection{\OurDataset Examples}
\label{app:qual-hat}

\begin{figure*}
    \centering
    \includegraphics[width=\linewidth]{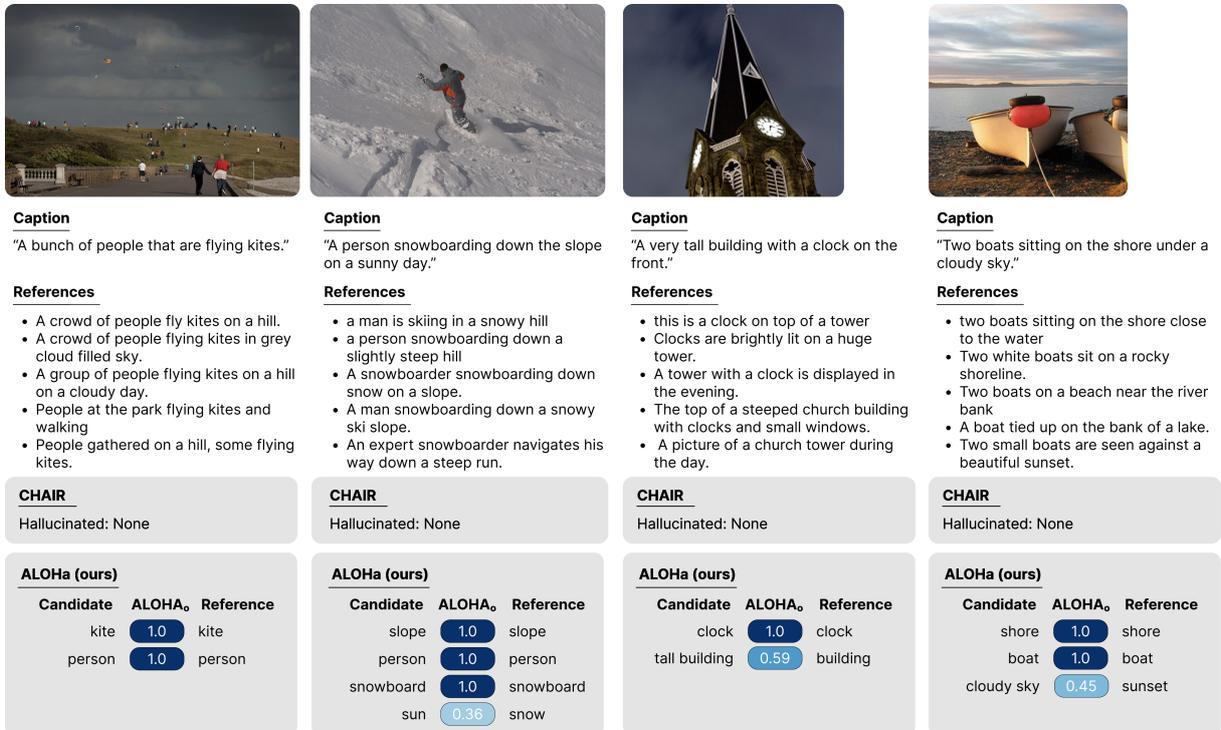}
    \caption{Randomly selected qualitative examples of \Ours evaluated on the \OurDataset dataset when there is no hallucination in the ground truth. See \autoref{app:qual-hat} for discussion.}
    \label{fig:app-hat-random-non-hc}
\end{figure*}

\begin{figure*}
    \centering
    \includegraphics[width=\linewidth]{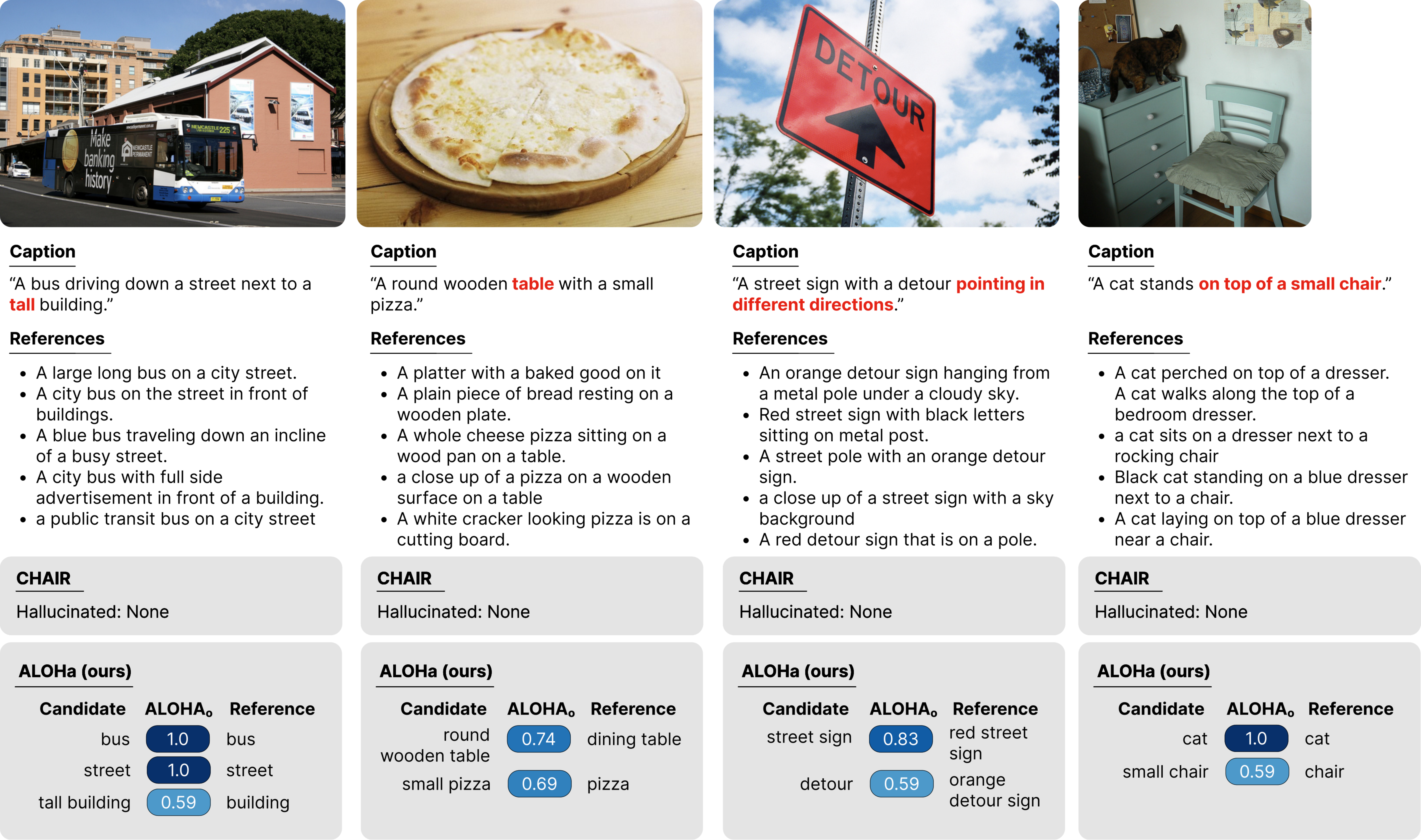}
    \caption{Randomly selected qualitative examples of \Ours evaluated on the \OurDataset dataset when there is a hallucination in the ground truth. These hallucinations are generally challenging to detect. See \autoref{app:qual-hat} for discussion.}
    \label{fig:app-hat-random-hc}
\end{figure*}

We present 4 random samples from \OurDataset each for cases without hallucinations (\autoref{fig:app-hat-random-non-hc}) and with hallucinations (\autoref{fig:app-hat-random-hc}). Because these examples contain more nuance than we discuss below, we do not indicate binary hallucination decisions as in~\autoref{app:qual-flickr}.

Starting with \autoref{fig:app-hat-random-non-hc}), samples with captions that were labeled as correct, from left to right:
{\RaggedRight
\begin{enumerate}
    \item Both CHAIR and \Ours successfully do not find any hallucinations.\\
    \item CHAIR does not flag any hallucinations. \Ours assigns a low \OScore$=0.36$ for ``sun``, an incorrect parse from the phrase ``sunny day''. However, the other objects are successfully matched. Interestingly, \Ours adds ``snowboard'' as an object, inferring that the physical item would need to be present given the verb ``snowboarding''. \\
    \item CHAIR again does not flag any hallucinations. \OScore for ``tall building'' is the mid-range 0.59, matched with the reference ``building'', indicating a somewhat uncertain attribute. This may be reasonable given the point of view in the image.
    \item CHAIR finds no hallucinations. ``Cloudy sky'' receives a somewhat low \OScore$=0.45$. Although this phrase is accurate given the image, this is a failure case in which the references are incomplete.
\end{enumerate}
}

Next, we discuss \autoref{fig:app-hat-random-hc}, showing samples that were labeled to contain a hallucination. Recall that labels capture \textit{all} types of caption errors, including those other than object hallucinations, to serve as a valuable source for research around general caption correctness. As a result, there exist non-object hallucinations in \OurDataset that are impossible for CHAIR or \Ours to localize. From left to right:

\begin{enumerate}
    \item The attribute ``tall'' is labeled as a hallucination, as the building next to the bus is only one story. Similar to sample 3 in \autoref{fig:app-hat-random-non-hc}, \OScore for ``tall building'' is somewhat uncertain at 0.59. Other objects are correctly grounded.
    \item The object ``table'' is a hallucinated, misclassified object; e.g., one reference opts for the more general ``wooden surface.'' However, the reference mentions a ``table'' that it is placed on, leading CHAIR to avoid considering it as a hallucination. For \Ours, this example shows one of the 2.97\% of cases (\autoref{tab:llm}) where \Ours hallucinates a reference object, ``dining table''. The candidate ``round wooden table'' is matched to it, with an erroneously high \OScore of 0.74.
    \item This sample contains a complex error, in which the arrow is not, in fact, ``pointing in different directions.'' This non-object hallucination is impossible for the object-specific CHAIR and \Ours to localize correctly. However, it demonstrates \Ours's capability to extract more complex attributes such as ``red street sign'' and ``orange detour sign.''
    \item The cat's location ``on top of a small chair'' is labeled as an error. CHAIR does not flag any hallucinations. \OScore for ``small chair'' is 0.59, yet both metrics cannot capture the specific relation.

\end{enumerate}

\end{document}